\documentclass{article} % For LaTeX2e
\usepackage{iclr2027_conference,times}

\usepackage{amsmath,amsfonts,bm}

\def\eqref#1{equation~\ref{#1}}
\def\1{\bm{1}}

\DeclareMathAlphabet{\mathsfit}{\encodingdefault}{\sfdefault}{m}{sl}
\SetMathAlphabet{\mathsfit}{bold}{\encodingdefault}{\sfdefault}{bx}{n}

\usepackage{amssymb}

\usepackage{url}
\usepackage{booktabs,graphicx}
\usepackage{multirow}
\usepackage[table]{xcolor}
\usepackage{booktabs}
\usepackage{multirow}
\usepackage{pifont}
\usepackage{xcolor}
\usepackage{makecell}
\usepackage{graphicx} % 引入插图宏包=
\usepackage{booktabs}
\usepackage{tabularx}
\usepackage{array}
\usepackage{amsmath}
\usepackage{wrapfig}

\iclrfinalcopy

\usepackage{hyperref}
\definecolor{eqpurple}{rgb}{0.9,0.6,0.9}
\hypersetup{
    colorlinks=true,
    citecolor=blue,         
    linkcolor=eqpurple,      
    urlcolor=eqpurple          
}
\definecolor{GSAPOrange1}{RGB}{255,246,235}
\definecolor{GSAPOrange2}{RGB}{255,239,220}
\definecolor{GSAPOrange3}{RGB}{255,229,199}
\definecolor{GSAPOrange4}{RGB}{253,217,177}
\definecolor{GSAPOrange5}{RGB}{250,200,150}
\definecolor{GSAPGray}{RGB}{245,245,245}

\title{Rethinking Pairwise Token Interaction in Spiking Transformers}

\author{Sicheng Shen$^{1,2,3,*}$,\quad Dongcheng Zhao$^{2,*}$,\quad Zhiyuan Li$^{2,3}$,\quad Jinyan Yu$^{2}$, \quad Qian Zhang$^{1}$\\ 
\textbf{Dengpeng Xing}$^{1,\dagger}$\textbf{,}\quad \textbf{Zhitong Zhang}$^{3,\dagger}$\textbf{,}\quad \textbf{Tielin Zhang}$^{2,\dagger}$ 
\\
\\
$^{1}$  Institute of Automatoin, CAS  \\
$^{2}$  Center for Excellence in Brain Science and Intelligence Technology,  State Key Laboratory of \\ \hspace*{0.45em} Brain Cognition and  Brain-inspired Intelligence Technology,Institute of Neuroscience, CAS \\
$^{3}$ Zhongguancun Academy \qquad $^*$ Equal Contribution \qquad $^\dagger$ Corresponding Author \\
\texttt{\small dengpeng.xing@ia.ac.cn, zhangzhitong@bza.edu.cn, zhangtielin@@ion.ac.cn} \\
}

\begin{document}

\maketitle

\begin{abstract}
Spiking Transformers inherit token interaction mechanisms from conventional Transformers, yet their sparse binary representations fundamentally alter how token-to-token communication is established. In particular, spike-based query–key matching produces highly sparse and input-dependent interaction patterns, coupling information propagation to the instantaneous availability of matching spike events. This motivates a different interaction paradigm in which long-range communication does not rely solely on pairwise spike coincidence.  
We therefore propose \textbf{Gated Spike Axial Propagation (GSAP)}, a spike-native token interaction mechanism that decouples information propagation from context selection. Instead of directly determining communication through query–key matching, GSAP first propagates spike-based context along the horizontal and vertical axes, allowing information to reach distant tokens through structured sequential propagation. A receiver-conditioned gate then determines how much of the propagated context is incorporated at each token, while a lightweight local pathway preserves fine-grained neighborhood information. In this way, GSAP reformulates token interaction as a propagate-then-select process, enabling structured long-range communication while retaining the sparse event-driven nature of spiking representations.
Code is available at \url{https://github.com/Fancyssc/GSAP}.
\end{abstract}

\section{Introduction}

Spiking Neural Networks (SNNs) provide a biologically inspired paradigm
for energy-efficient computation by representing information through sparse
spike events and temporally evolving neuronal states
~\citep{maass1997networks,gilra2017predicting}.
Their event-driven nature offers promising opportunities for neuromorphic
hardware and resource-constrained intelligent systems
~\citep{roy2019towards}.
However, the discontinuous dynamics of spike generation introduce
significant optimization challenges for deep SNNs. Recent advances in
surrogate-gradient learning, learnable neuronal dynamics, and residual
spiking architectures have substantially improved the trainability and
scalability of SNNs
~\citep{neftci2019surrogate,fang2021incorporating,
fang2021deep,eshraghian2023training},
enabling increasingly expressive spiking models for visual recognition.

Transformers achieve powerful representation capability by modeling
long-range dependencies through dynamic token interaction
~\citep{vaswani2017attention,dosovitskiy2020image}.
To combine the interaction capability of Transformers with spike-driven
computation, Spikformer introduced Spiking Self-Attention (SSA), which
adapts self-attention to binary spike representations by removing Softmax
and operating on spike-based queries, keys, and values
~\citep{zhou2022spikformer}.
Following this direction, subsequent studies have explored improved
spike-driven attention mechanisms, hierarchical architectures, temporal
representation learning, and efficient deployment strategies
~\citep{yao2023spike,yao2024spike,zhou2026spikingformer,
zhou2024qkformer,shi2024spikingresformer,shen2024tim,
lee2025spiking,liu2024sparsespikformer}.
These advances have established Spiking Transformers as a promising
architecture family that combines Transformer-style representation
learning with event-driven computation.

Despite these advances, existing Spiking Transformers largely inherit the
token interaction principle of conventional Transformers, where
information exchange is established through explicit pairwise
relationships between tokens. While effective for dense-valued
representations, this interaction paradigm introduces a potential
mismatch with sparse spike-based representations. In Spiking Attention,
the contribution of a source token to a receiving token is determined by
the coincidence between their query and key spike patterns. Therefore,
communication pathways are explicitly constructed through discrete
matching events rather than continuously available interaction patterns.
Under sparse spike activity, effective information exchange becomes
dependent on whether compatible spike patterns exist between source and
receiver tokens~\citep{guo2025spiking,xiao2025rethinking,zhao2026spiking}.

\begin{figure}[t]
    \centering
    \includegraphics[width=1\linewidth]{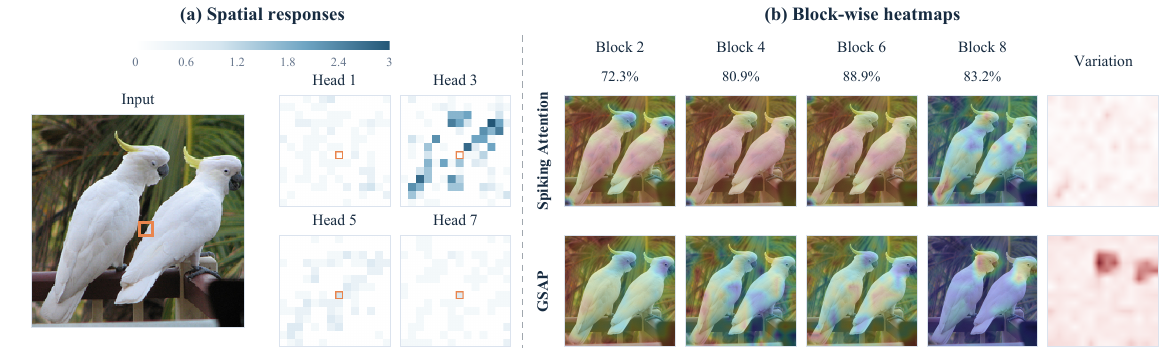}
    \caption{\small 
(a) Spatial response maps for selected heads of Spiking Self-Attention
(SSA) in an ImageNet-1K 8-768 model with $T=4$. 
(b) Block-wise response visualizations for SSA and GSAP. 
The percentages annotate the fractions of tokens without Q--K matches over
four time steps for the displayed example. 
\textit{Variation} visualizes changes across blocks.}
\label{fig:sparsity}
\end{figure}

This observation motivates revisiting token interaction in Spiking
Transformers from the perspective of spike-based communication. Instead
of explicitly constructing source--receiver affinities, we argue that
token interaction can be decomposed into two complementary functions:
\textit{global context interaction} and
\textit{receiver-conditioned selection}. The former determines how
information is exchanged across spatial locations, whereas the latter
controls how each token utilizes the received contextual information
according to its current state.

Figure \ref{fig:sparsity} illustrates the sparse interaction patterns observed in SSA. The visualization shows that token communication is strongly influenced by discrete spike matching events, thereby impairing the ability to capture global dependencies and motivating the exploration of alternative interaction mechanisms beyond explicit pairwise affinity construction. Based on this perspective, we propose\textbf{ Gated Spike Axial Propagation (GSAP)}, a spike-native token interaction mechanism for Spiking Transformers. GSAP replaces explicit QKV matching with spike-driven global context interaction and receiver-conditioned selection. By decoupling communication from context selection, GSAP enables flexible long-range information exchange while preserving the temporal characteristics of spiking computation.

We evaluate GSAP on static-image classification, event-based recognition,
and semantic segmentation tasks using multiple Spiking Transformer
backbones. Experimental results demonstrate consistent improvements over
existing interaction mechanisms while maintaining competitive
computational efficiency. Extensive ablation studies further analyze the
roles of global context interaction and receiver-conditioned selection,
providing insights into how spike-driven token interaction can be
redesigned. Our contributions are summarized as follows:

\begin{itemize}

\item We identify a mismatch between pairwise query--key matching and sparse
spike representations: source-to-receiver communication relies on discrete
spike coincidences, limiting the interactions available for global feature
modeling.

\item We propose \textbf{GSAP}, a spike-native interaction module that
decouples spatial communication from receiver-conditioned selection.
Sequential horizontal--vertical spike propagation builds long-range structural
paths, while a receiver-side gate selects global context before fusion with
local and identity features.

\item Token-perturbation analyses reveal long-range communication through sequential axial propagation alongside spatially concentrated output sensitivity. Component ablations quantify the contributions of axial propagation, receiver-conditioned selection, and local refinement to recognition performance.

\end{itemize}
\section{Related Work}

\subsection{Spiking Transformer Architectures}

Spiking Transformers aim to combine the representation capability of
Transformers with the sparse and event-driven computation of SNNs.
Spikformer first established this paradigm by introducing Spiking
Self-Attention (SSA) and spike-based Transformer representations
~\citep{zhou2022spikformer}. Subsequent studies investigated more
spike-compatible architectures by redesigning computational pathways,
residual structures, and feature extraction mechanisms.
Spike-driven Transformer and Spikingformer reduced non-spike operations
and improved spike-driven execution, while Meta-SpikeFormer further
extended Spiking Transformers toward general-purpose visual backbones
~\citep{yao2023spike,yao2024spike}.

Beyond single-stage architectures, recent works explored hierarchical and
multi-scale Spiking Transformers. Spikformer V2 introduced a spiking
convolutional stem for improved visual representation, while QKFormer and
SpikingResformer developed hierarchical designs with multi-stage feature
processing~\citep{zhou2024qkformer,shi2024spikingresformer}. Meanwhile,
Spiking Transformers have been extended to dense prediction and other
vision tasks, including semantic segmentation and point-cloud
understanding~\citep{lei2025spike2former,wu2025spiking,lu2026spiking}.

These studies mainly focus on improving backbone scalability,
representation capability, and task generalization. In contrast, we
investigate the fundamental token interaction mechanism within Spiking
Transformers.

\subsection{Token Interaction in Spiking Transformers}

Token interaction is a core component of Transformer architectures.
Existing Spiking Transformers generally inherit attention-based
interaction from conventional Transformers and adapt it to binary spike
representations. Spikformer introduced SSA by replacing continuous
attention computation with spike-form query, key, and value representations
~\citep{zhou2022spikformer}. Subsequent approaches further enhanced
spatial and temporal interaction through improved attention structures,
including temporal-aware interaction mechanisms and spatiotemporal
attention designs~\citep{shen2024tim,lee2025spiking,lu2025estsformer}.

However, the sparse nature of spike representations may weaken the
relational information captured by explicit query--key matching. Recent
works therefore attempt to improve spike-based attention by enriching or
recovering token relationships. $\alpha$-XNOR incorporates non-spike
matches into binary similarity computation to alleviate information loss
~\citep{xiao2025rethinking}; SLI-ACF introduces additional local pathways
to recover neighborhood interactions~\citep{zhao2026spiking}; and HAST and
SQKformer enhance attention through composite tokens, channel-wise
interaction, or multi-path modeling
~\citep{fan2025hybrid,chen2026sqkformer}.

Different from these approaches, which preserve the fundamental
source--receiver matching formulation and improve its effectiveness, GSAP
revisits the interaction principle itself. Instead of explicitly computing
pairwise affinities, GSAP establishes token communication through spike
propagation and performs receiver-conditioned context selection.

\begin{figure}[t]
    \centering
    \includegraphics[width=\textwidth]{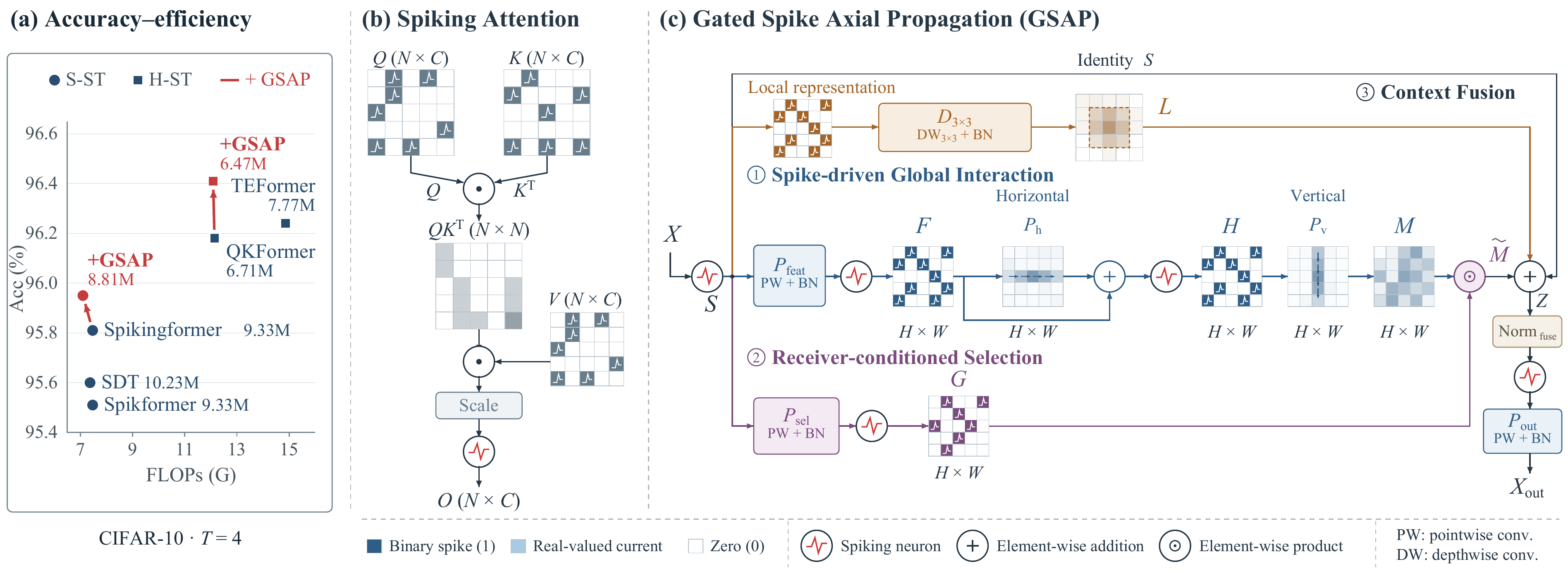}
    \caption{\small Overview of GSAP as a spike-native token interaction
    mechanism for Spiking Transformers. 
    (a) Accuracy--efficiency comparison with existing approaches.
    (b) Existing Spiking Attention establishes token interaction through
    explicit pairwise query--key matching.
    (c) GSAP replaces pairwise matching with propagation-based spike
    communication and receiver-conditioned context selection.}
    \label{fig:pipe}
\end{figure}

\section{Proposed Method}
\label{sec:method}

% We introduce \textbf{Gated Spike Axial Propagation (GSAP)}, 
% a spike-native token interaction mechanism for visual Spiking Transformers.
% Different from existing approaches that inherit explicit pairwise token
% matching from conventional Transformers, GSAP reformulates token
% interaction as a communication--selection process. Specifically, GSAP
% first establishes global contextual interaction through spike-driven
% spatial aggregation, then determines how each receiver integrates the
% aggregated information according to its current state. Finally, local
% structures and receiver-adapted global context are fused to update token
% representations. This design decouples communication from context
% selection while exploiting the temporal dynamics of spiking neurons.

We introduce \textbf{Gated Spike Axial Propagation (GSAP)}, a spike-native token interaction mechanism for visual Spiking Transformers. GSAP separates spatial communication from receiver-conditioned context selection: sequential horizontal and vertical propagation aggregates information across spatial locations, while a receiver-side gate regulates the utilization of the aggregated context. Spiking neurons maintain temporal membrane states within these computations, connecting spatial interaction with temporal processing. As illustrated in Fig. 2(c), the selected context is fused with local features and the input spikes to update token representations. We first review the neuronal dynamics and attention formulation that motivate this design, then describe the individual GSAP components.

\subsection{Background and Motivation: Spiking Transformer}
\label{sec:motivation}

Spiking Transformers combine the temporal dynamics of spiking neurons with
the token interaction capability of Transformers. A typical Spiking
Transformer block mainly consists of spiking neuron dynamics and
spike-driven token interaction. We briefly introduce these components and
discuss the limitation arising from their combination.

\paragraph{Spiking Neuron Dynamics.}

Spiking neurons provide the temporal computation mechanism in SNNs.
Following the leaky integrate-and-fire (LIF) model, the membrane state $U[t]$ and
spike generation $S[t]$ are formulated as

\begin{equation}
\begin{aligned}
U[t]
&=
\lambda U[t-1]+I[t],
\\
S[t]
&=
H(U[t]-V_{\mathrm{th}}),
\end{aligned}
\label{eq:lif}
\end{equation}

where $U[t]$ denotes the membrane potential, $I[t]$ is the input current,
$\lambda$ represents the leakage factor, and $H(\cdot)$ denotes the
Heaviside function. The membrane state provides temporal memory across
simulation steps and enables spike-driven information processing.

\paragraph{Spiking Self-Attention.}

To introduce Transformer-style token interaction into SNNs,
Spikformer proposed Spiking Self-Attention (SSA), where query, key, and
value representations are encoded as binary spike sequences.
For one attention head at time step $t$, SSA performs

\begin{equation}
\mathbf{J}^{t}
=
s\left(\mathbf{Q}^{t}(\mathbf{K}^{t})^\top\right)\mathbf{V}^{t},
\label{eq:sa_matching}
\end{equation}

where $\mathbf{Q}^{t},\mathbf{K}^{t},\mathbf{V}^{t}\in\{0,1\}^{N\times d}$,
$N$ denotes the number of tokens, and $d$ is the head dimension.
The interaction strength between tokens is determined by query--key spike
coincidence, which explicitly constructs communication relationships
before context aggregation.

Although SSA enables content-dependent token interaction, it couples
communication and context selection through sparse spike matching.
GSAP instead decouples these two functions by using spike-driven global
context propagation for communication and receiver-conditioned selection
for adaptive context utilization.

\subsection{Gated Spike Axial Propagation}
\label{sec:gsap}
As shown in Fig. 2(c), GSAP processes the input spike representation through parallel propagation, selection, and local pathways. The propagation pathway projects the input into spike features and performs horizontal aggregation, spike re-encoding, and vertical aggregation in sequence to obtain the contextual representation \(M\). The selection pathway generates a binary gate \(G\) from the receiver-side input and its neuronal state, yielding the selected context \(\widetilde{M}=G\odot M\). The local pathway extracts neighborhood features \(L\), which are added to \(\widetilde{M}\) and the identity input \(S\). Normalization, spike re-encoding, and output projection then produce the updated representation. The following paragraphs formalize these computations.

\paragraph{Spike-driven Global Interaction.}

GSAP establishes long-range token interaction through spike-driven global
context aggregation. Given an input spike representation
$\mathbf{S}\in\{0,1\}^{T\times B\times C\times H\times W}$,
GSAP first projects the input spikes into spike features and subsequently
aggregates contextual information along two orthogonal spatial directions:

\begin{equation}
\begin{aligned}
\mathbf{F}
&=
\mathrm{SN}_{\mathrm{feat}}
\left(
\mathcal{P}_{\mathrm{feat}}(\mathbf{S})
\right),
\\
\mathbf{H}
&=
\mathrm{SN}_{h}
\left(
\mathcal{P}_{h}(\mathbf{F})
+
\mathbf{F}
\right),
\\
\mathbf{M}
&=
\mathcal{P}_{v}(\mathbf{H}),
\end{aligned}
\label{eq:gsap_global}
\end{equation}

where $\mathcal{P}_{\mathrm{feat}}(\cdot)$ denotes a pointwise projection,
and $\mathcal{P}_{h}(\cdot)$ and $\mathcal{P}_{v}(\cdot)$ represent
horizontal and vertical spatial aggregation operators, respectively. The
intermediate spiking neuron re-encodes the horizontally aggregated
responses and preserves temporal membrane states, allowing global context
aggregation to be modulated by spike-driven temporal dynamics. Through sequential axial aggregation, information from distant spatial
locations can contribute to each receiving token without explicitly
constructing source--receiver affinities. Different from attention, where
the interaction topology is determined by query--key similarity before
aggregation, GSAP establishes token communication through learned spatial
aggregation dynamics and neuronal temporal states.

\paragraph{Receiver-conditioned Selection.}

After global context aggregation, the received information does not
necessarily contribute equally to every token. GSAP therefore introduces
a receiver-conditioned selection mechanism, where each token determines
how the aggregated context should be integrated according to its own
representation and temporal neuronal state.

A parallel branch generates a receiver-dependent selection signal:

\begin{equation}
\begin{aligned}
\mathbf{G}
&=
\mathrm{SN}_{\mathrm{sel}}
\left(
\mathcal{P}_{\mathrm{sel}}(\mathbf{S})
\right),
\\
\widetilde{\mathbf{M}}
&=
\mathbf{G}
\odot
\mathbf{M},
\end{aligned}
\label{eq:gsap_selection}
\end{equation}

where $\mathcal{P}_{\mathrm{sel}}(\cdot)$ denotes a lightweight projection,
$\mathrm{SN}_{\mathrm{sel}}$ represents a spiking neuron, and
$\odot$ denotes element-wise multiplication. The selection signal has the
same spatial and channel dimensions as the aggregated context $\mathbf{M}$.
Unlike source-dependent weighting mechanisms, the selection signal is
generated from the receiver-side representation, allowing each token to
regulate context utilization according to its current state.

Unlike attention-based interaction, where source--receiver relationships
and context selection are jointly determined by affinity estimation,
GSAP decouples communication from selection.
Specifically, global propagation enables information exchange across tokens,
while receiver-conditioned selection adaptively regulates the utilization
of received context according to each token's representation and temporal
state.

\paragraph{Context Fusion.}

While global context interaction enables long-range information exchange,
local spatial structures remain important for visual representation
learning. GSAP therefore preserves local information through a lightweight
local representation pathway:

\begin{equation}
\begin{aligned}
\mathbf{L}
&=
\mathcal{D}_{3\times3}(\mathbf{S}),
\\
\mathbf{Z}
&=
\mathbf{S}
+
\mathbf{L}
+
\widetilde{\mathbf{M}},
\\
\mathbf{X}_{out}
&=
\mathcal{P}_{\mathrm{out}}
\left(
\mathrm{SN}_{\mathrm{fuse}}
\left(
\mathrm{Norm}_{\mathrm{fuse}}
(\mathbf{Z})
\right)
\right).
\end{aligned}
\label{eq:gsap_fusion}
\end{equation}

where $\mathcal{D}_{3\times3}$ denotes a depthwise convolution operating
on spike representations and $\mathrm{Norm}_{\mathrm{fuse}}$ denotes the
normalization operation before spike re-encoding. The fused
representation combines the original spike state, local spatial
structures, and receiver-selected global context.

Through this fusion process, GSAP integrates local refinement and global
context propagation to generate updated token representations.
Different from attention-based interaction that relies on explicit
pairwise affinity estimation, GSAP forms implicit communication pathways
through spike-driven propagation and receiver-conditioned message
selection.
\section{Experiments}
\label{sec:experiments}

We evaluate GSAP as a spike-native token interaction mechanism for
Spiking Transformers across different architectures and tasks.
Specifically, we conduct experiments on image classification,
event-based recognition, and semantic segmentation.
In all experiments, GSAP is integrated by replacing the original token
interaction modules while keeping the remaining backbone components
unchanged, ensuring a fair comparison with existing Spiking Transformer
architectures.
Dataset, implementation, and training details are provided in Appendices \ref{ap:imagenet} and \ref{ap:seg}, with energy and computational complexity analysis in Appendix \ref{ap:complexity}.

% \subsection{Main Results}
% \label{sec:main_results}

% We first evaluate whether GSAP provides an effective alternative to
% attention-based token interaction. GSAP is integrated into both
% single-stage and hierarchical Spiking Transformer architectures, while
% keeping the remaining backbone components unchanged.

\begin{table*}[!b]
\vspace{-6mm}
\centering
\caption{\small ImageNet-1K classification results.
$^*$ denotes an input resolution of $288\times288$; all other models
use $224\times224$. GSAP achieves significantly better performance than the baseline across nearly all model sizes while using fewer parameters.
}
\label{tab:imagenet}
\resizebox{0.85\textwidth}{!}{
\begin{tabular}{cccccc}
\toprule
\textbf{Methods} & \textbf{Type} & \textbf{Architecture} & \makecell[c]{\textbf{Params}\\\textbf{(M)}} & \makecell[c]{\textbf{Time}\\\textbf{Steps}} & \textbf{Acc.}\textbf{(\%)} \\
\midrule

ResNet~\citep{he2016deep}
& ANN & Res-CNN-104 & 54.21 & 1 & 76.87 \\

\cmidrule{2-6}

ViT~\citep{dosovitskiy2020image}
& ANN & ViT-B/16 & 86.59 & 1 & 77.90 \\

\cmidrule{2-6}

\multirow{2}{*}{DeiT~\citep{touvron2021training}}
& \multirow{2}{*}{ANN} & DeiT-B & 86.59 & 1 & 81.80 \\
& & DeiT-B$^*$ & 86.59 & 1 & 83.10 \\

\midrule

\multirow{2}{*}{Spiking ResNet~\citep{hu2021spiking}}
& \multirow{2}{*}{SNN} & Spiking-ResNet-34  & 21.79 & 350 & 71.61 \\
& & Spiking-ResNet-51 & 25.56 & 350 & 72.75 \\

\cmidrule{2-6}

\multirow{3}{*}{SEW ResNet~\citep{fang2021deep}}
& \multirow{3}{*}{SNN} & SEW-ResNet-34  & 21.79 & 4 & 67.04 \\
& & SEW-ResNet-101 & 44.55 & 4 & 68.76 \\
& & SEW-ResNet-152 & 60.19 & 4 & 69.26 \\

\cmidrule{2-6}

\multirow{3}{*}{Att MS-ResNet~\citep{yao2022attention}}
& \multirow{3}{*}{SNN} & Att-MS-ResNet-18  & 11.69 & 6 & 63.10 \\
& & Att-MS-ResNet-34 & 21.80 & 6 & 69.42 \\
& & Att-MS-ResNet-104 & 77.28 & 5 & 76.02 \\

\midrule

\multirow{3}{*}{Spikformer~\citep{zhou2022spikformer}}
& \multirow{3}{*}{SNN} & Spikformer-8-384 & 16.81 & 4 & 70.24 \\
&  & Spikformer-8-512 & 29.68 & 4 & 73.38 \\
&  & Spikformer-8-768 & 66.34 & 4 & 74.81 \\

\cmidrule{2-6}

\multirow{3}{*}{SDT~\citep{yao2023spike}}
& \multirow{3}{*}{SNN} & SDT-8-384 & 16.81 & 4 & 72.28 \\
& & SDT-8-512 & 29.68 & 4 & 74.57 \\
&  & SDT-8-768$^*$ & 66.34 & 4 & 77.07 \\

\cmidrule{2-6}

\multirow{3}{*}{Spikingformer~\citep{zhou2026spikingformer}}
& \multirow{3}{*}{SNN} & Spikingformer-8-384 & 16.81 & 4 & 72.45 \\
& & Spikingformer-8-512 & 29.68 & 4 & 74.79 \\
& & Spikingformer-8-768 & 66.34 & 4 & 75.85 \\

\cmidrule{2-6}

\multirow{2}{*}{$\alpha$-SSA~\citep{xiao2025rethinking}}
& \multirow{2}{*}{SNN} & $\alpha$-SSA-ViT-8-384 & 16.80 & 4 & 71.36 \\
& & $\alpha$-SSA-ViT-8-512 & 29.70 & 4 & 75.39 \\

\cmidrule{2-6}

\multirow{2}{*}{SEMM ~\citep{zhou2024spiking}}
& \multirow{2}{*}{SNN} & Spikingformer-8-384 & 16.05 & 4 & 73.58 \\
& & Spikingformer-8-512 & 28.22 & 4 & 76.03 \\

\midrule

\multirow{3}{*}{\textbf{GSAP(ours)}}
& \multirow{3}{*}{SNN} & Spikingformer-8-384 & 15.85 & 4 & \textbf{73.29}\\
& & Spikingformer-8-512 & 27.87 & 4 & \textbf{76.07}\\
& & Spikingformer-8-768 & 62.04 & 4 &  \textbf{78.19} \\

\bottomrule
\end{tabular}
}
\end{table*}

\subsection{Image Classification}
\label{sec:image_classification}

We first evaluate GSAP on image classification tasks, which serve as the standard benchmark for Spiking Transformers. Experiments are conducted on ImageNet-1K, CIFAR-10, CIFAR-100, and TinyImageNet to investigate the effectiveness of GSAP under different model scales and architectural configurations~\citep{deng2009imagenet,krizhevsky2010convolutional}.

% We evaluate GSAP across static-image classification, event-based recognition, and semantic segmentation to assess its effectiveness across tasks and backbone architectures. Classification experiments report accuracy and parameter counts, while segmentation reports mIoU, mAcc, and aAcc. TinyImageNet and UCF101-DVS use the unified configurations specified in the respective table captions. Propagation analyses and component ablations connect task performance to the module's design. Further experiments evaluate alternative input encodings and online learning, with the latter using an ungated GSAP variant.
As shown in Table~\ref{tab:imagenet}, GSAP consistently improves the performance of Spiking Transformer backbones across different model scales. Specifically, GSAP improves Spikingformer-8-384, Spikingformer-8-512, and Spikingformer-8-768 from 72.45\%, 74.79\%, and 75.85\% to 73.29\%, 76.07\%, and 78.19\%, respectively. Meanwhile, GSAP reduces the corresponding parameter counts from 16.81M, 29.68M, and 66.34M to 15.85M, 27.87M, and 62.04M. Interestingly, the performance gain becomes more pronounced with larger model scales, suggesting that propagation-based interaction can better exploit large-scale token communication. These results demonstrate that effective token interaction in Spiking Transformers can be achieved through propagation-based communication without introducing additional parameters. 

To further investigate whether GSAP is coupled with a specific Spiking Transformer architecture, we evaluate it on both single-stage and hierarchical Spiking Transformer families. As shown in Table~\ref{tab:cifar}, GSAP consistently improves different architectural variants under comparable settings. For single-stage Spiking Transformers, GSAP improves Spikingformer from 79.21\% to 80.21\% on CIFAR-100 while reducing the parameter count from 9.36M to 8.84M. For hierarchical Spiking Transformers, GSAP improves QKFormer from 81.15\% to 81.64\% with fewer parameters. Similar improvements are observed on TinyImageNet, where GSAP achieves 66.84\% and 66.65\% for single-stage and hierarchical architectures, respectively. 

Overall, these results demonstrate that GSAP provides a general spike-native token interaction mechanism that is not tailored to a specific backbone, but can be effectively applied to diverse Spiking Transformer architectures.

\begin{table*}[t]
    \vspace{-5mm}
    \centering
\caption{\small
Classification results on CIFAR-10, CIFAR-100, and TinyImageNet.
S-ST denotes single-stage Spiking Transformer, while H-ST denotes
hierarchical Spiking Transformer.
}
    \label{tab:cifar}
    \label{tab:classification_results}
    \renewcommand{\arraystretch}{1.25}
    \resizebox{\textwidth}{!}{
    \begin{tabular}{cc ccc ccc ccc}
        \toprule

        \multirow{2}{*}{\textbf{Method}}
        & \multirow{2}{*}{\textbf{Architecture}}
        & \multicolumn{3}{c}{\textbf{CIFAR10}}
        & \multicolumn{3}{c}{\textbf{CIFAR100}}
        & \multicolumn{3}{c}{\textbf{TinyImageNet}$^{\dagger}$} \\

        \cmidrule(lr){3-5}
        \cmidrule(lr){6-8}
        \cmidrule(lr){9-11}

        &
        & Params (M) & $T$ & Acc. (\%)
        & Params (M) & $T$ & Acc. (\%)
        & Params (M) & $T$ & Acc. (\%) \\
        \midrule

        Spikformer~\citep{zhou2022spikformer} & S-ST
        & 9.33 & 4 & 95.51
        & 9.37 & 4 & 78.21
        & 9.40 & 4 & 64.78 \\

        SDT~\citep{yao2023spike} & S-ST
        & 10.23 & 4 & 95.60
        & 10.28 & 4 & 78.40
        & 9.39 & 4 & 66.03 \\

        RevSFormer~\citep{zhang2024memory} & S-ST
        & 9.33 & 4 & 95.34
        & 9.37 & 4 & 79.04
        & - & - & - \\

        STAttn+Spikf~\citep{lee2025spiking} & S-ST
        & 9.33 & 4 & 94.36
        & - & 4 & 75.85
        & - & - & - \\

        Spikingformer~\citep{zhou2026spikingformer} & S-ST
        & 9.33 & 4 & 95.81
        & 9.36 & 4 & 79.21
        & 9.39 & 4 & 65.13 \\

        \rowcolor{GSAPGray}\textbf{GSAP (ours)} & S-ST
        & 8.81 & 4 & \cellcolor{GSAPOrange1}\textbf{95.95}
        & 8.84 & 4 & \cellcolor{GSAPOrange4}\textbf{80.21}
        & 8.92 & 4 & \cellcolor{GSAPOrange3}\textbf{66.84} \\

        \midrule

        QKFormer~\citep{zhou2024qkformer} & H-ST
        & 6.71 & 4 & 96.18
        & 6.74 & 4 & 81.15
        & 6.78 & 4 & 64.07 \\

        STAttn+QKF~\citep{lee2025spiking}  & H-ST
        & 6.71 & 4 & 95.35
        & 6.74 & 4 & 80.20
        & - & - & - \\

TEFormer~\citep{shen2026teformer} & H-ST
        & 7.77 & 4 & 96.24
        & 7.80 & 4 & 79.84
        & - & - & - \\

        \rowcolor{GSAPGray}\textbf{GSAP (ours)} & H-ST
        & 6.47 & 4 & \cellcolor{GSAPOrange1}\textbf{96.41}
        & 6.51 & 4 & \cellcolor{GSAPOrange2}\textbf{81.64}
        & 6.60 & 4 & \cellcolor{GSAPOrange5}\textbf{66.65} \\

        \bottomrule
    \end{tabular}
    }
\end{table*}
% \subsection{Event-Based Recognition}

\subsection{Event-based Datasets}
\label{sec:event_recognition}

Beyond frame-based image inputs, we further evaluate GSAP on event-based tasks to investigate its compatibility with sparse and
temporally structured spike-driven data.
Experiments are conducted on CIFAR10-DVS, N-Caltech101, and UCF101-DVS~\citep{li2017cifar10,orchard2015converting,bi2020graph}.

As shown in Table~\ref{tab:dvs}, replacing the original token interaction
modules with GSAP consistently improves recognition performance across
different event-based datasets.
In particular, GSAP improves the accuracy on N-Caltech101 from 84.45\% to
85.54\%.
This result indicates that propagation-based interaction is effective for
processing sparse temporal event streams, where information is naturally
distributed across time and accumulated through dynamic interactions.
These observations further support the motivation of GSAP as a
spike-native interaction mechanism.

\begin{table*}[h]
    \vspace{-3mm}
    \centering
\caption{\small
Classification results on CIFAR10-DVS, N-Caltech101, and UCF101-DVS.
}
    \label{tab:dvs}
    \label{tab:dvs_classification_results}
    \renewcommand{\arraystretch}{1.2}
    \resizebox{\textwidth}{!}{
    \begin{tabular}{c ccc ccc ccc}
        \toprule
        \multirow{2}{*}{\textbf{Method}}
        & \multicolumn{3}{c}{\textbf{CIFAR10-DVS}}
        & \multicolumn{3}{c}{\textbf{N-Caltech101}}
        & \multicolumn{3}{c}{\textbf{UCF101-DVS}$^{\dagger}$} \\
        \cmidrule(lr){2-4}
        \cmidrule(lr){5-7}
        \cmidrule(lr){8-10}
        & Params (M) & $T$ & Acc. (\%)
        & Params (M) & $T$ & Acc. (\%)
        & Params (M) & $T$ & Acc. (\%) \\
        \midrule
        SALT~\citep{kim2021optimizing}
        & - & 20 & 67.1
        & - & 20 & 55.0
        & - & - & -\\
        Res-SNN18+RM~\citep{yao2023sparser}
        & - & - & -
        & - & 30 × 10 & 81.2
        & - & 8 & 58.5 \\
        NDA~\citep{li2022neuromorphic}
        & - & 10 & 81.5
        & - & 10 & 79.5
        & - & - & - \\
        EventMix+ResNet~\citep{shen2023eventmix}
        & - & 10 & 78.0
        & - & 10 & 78.6
        & - & - & - \\
        \midrule
        Spikformer~\citep{zhou2022spikformer}
        & 2.59 & 16 & 80.60
        & 2.61 & 10 & 83.96
        & 2.61 & 10 & 64.83 \\
        SDT~\citep{yao2023spike}
        & 2.57 & 16 & 80.00
        & 2.59 & 10 & 83.23
        & 2.59 & 10 & 65.23 \\
        TIM~\citep{shen2024tim}
        & 2.57 & 16 & 81.60
        & 2.59 & 10 & 79.00
        & 2.59 & 10 & 63.80 \\
        Spikingformer~\citep{zhou2026spikingformer}
        & 2.57 & 16 & 81.30
        & 2.59 & 10 & 84.45
        & 2.59 & 10 & 65.56 \\
        \rowcolor{GSAPGray} \textbf{GSAP (ours)}
        & 2.46 & 16 & \cellcolor{GSAPOrange1}\textbf{81.80}
        & 2.48 & 10 & \cellcolor{GSAPOrange3}\textbf{85.54}
        & 2.48 & 10 & \cellcolor{GSAPOrange1}\textbf{65.64}\\
        \bottomrule
    \end{tabular}
    }
\end{table*}

% We evaluate GSAP on CIFAR10-DVS, N-Caltech101, and UCF101-DVS~\citep{li2017cifar10,orchard2015converting,bi2020graph}. Table~\ref{tab:dvs} reports accuracy gains of 0.50, 1.09, and 0.08 percentage points over Spikingformer, respectively, with fewer parameters in each configuration. The largest gain occurs on N-Caltech101, while the UCF101-DVS difference is small. These results extend the evidence for GSAP's spatial interaction design to event-based object and action recognition. Alternative input-encoding results are reported in Appendix~\ref{ap:neuron encoding}.

% \subsection{Semantic Segmentation}

% We evaluate dense prediction on ADE20K and PASCAL VOC~\citep{zhou2017scene,everingham2010pascal}. Table~\ref{tab:seg} reports ADE20K mIoU improvements of 4.27 and 2.18 percentage points for the QKFormer-based and Spikingformer-based adaptations, respectively, together with improvements in mAcc and aAcc. On PASCAL VOC, the QKFormer-based adaptation gains 1.47 points in mIoU and 0.69 points in the reported mAcc, while the reported aAcc decreases by 1.14 points. The consistent mIoU gains extend the evaluation to dense prediction. Architecture details and selected qualitative examples are provided in Appendix~\ref{ap:seg}.
\subsection{Semantic Segmentation}
\label{sec:semantic_segmentation}

\begin{wraptable}{r}{0.45\linewidth}
    \centering
    \vspace{-4mm}
    \caption{\small Semantic segmentation results on ADE20K and PASCAL VOC. 
    Each +GSAP row replaces the interaction modules of the preceding baseline.}
    \label{tab:seg}
    \resizebox{0.85\linewidth}{!}{
    \begin{tabular}{c c c c c c}
    \toprule
    \textbf{Model} & \textbf{Dataset} &\textbf {mIoU} & \textbf{mAcc} & \textbf{aAcc}\\
    \midrule
    QKFormer & \multirow{4}{*}{\textbf{ADE20K}} & 32.63 & 43.33 & 75.56\\
    \textbf{+GSAP} & & \textbf{36.90} & 47.88 & 77.63\\
    \cmidrule{1-1} \cmidrule{3-5}
    Spikingformer & & 31.64 & 41.85 & 74.99\\
    \textbf{+GSAP} & & \textbf{33.82} & 43.63 & 76.49\\
    \midrule
    QKFormer & \multirow{2}{*}{\textbf{VOC}} & 60.76 & 90.44 & 73.73\\
    \textbf{+GSAP} & & \textbf{62.23} & 91.13 & 72.59\\
    \bottomrule
    \end{tabular}
    }
\end{wraptable}

We further investigate whether GSAP can benefit dense prediction tasks,
where effective spatial information exchange among tokens is essential for
preserving fine-grained structures.

As shown in Table~\ref{tab:seg}, integrating GSAP into different Spiking
Transformer backbones consistently improves semantic segmentation
performance on ADE20K and PASCAL VOC~\citep{zhou2017scene,everingham2010pascal}.
On ADE20K, GSAP improves QKFormer from 32.63\% to 36.90\% mIoU and
improves Spikingformer from 31.64\% to 33.82\% mIoU.
On PASCAL VOC, GSAP also achieves improvements over the corresponding
baseline.

These results demonstrate that GSAP provides a generalizable spike-native
token interaction mechanism across different architectures, input
modalities, and prediction tasks.
\section{Analysis \& Discussion}
\label{sec:analysis}

% \subsection{Design Analysis of GSAP}
\subsection{Ablation Study}
\label{sec:ablation}
\begin{figure}[h]
    \centering
    \includegraphics[width=0.85\linewidth]{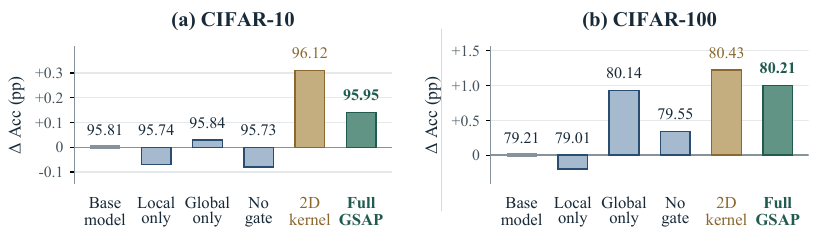}
\caption{\small
Ablation studies of GSAP on CIFAR-10 and CIFAR-100.
Component ablations evaluate global propagation, receiver gating, and local
refinement, while propagation-path ablations compare axial propagation with
full two-dimensional propagation.
The latter replaces both axial operators and the intermediate LIF neuron
and row residual connection.}
    \label{fig:ablation}
\end{figure}
We conduct ablation studies on CIFAR-10 and CIFAR-100 using the
Spikingformer backbone with $T=4$ to investigate the contribution of
different design choices in GSAP, including long-range propagation,
receiver-conditioned message selection, and propagation structures.

Starting from the full GSAP model, we progressively remove
components to analyze their effects.
The complete model achieves 95.95\% and 80.21\% accuracy on CIFAR-10 and
CIFAR-100, respectively.
Removing the global propagation branch decreases accuracy by 0.21 and
1.20 percentage points, demonstrating the importance of long-range message
aggregation.
Removing the receiver-conditioned gate further leads to accuracy drops of
0.22 and 0.66 percentage points, even when both local and global
propagation pathways remain available.
This indicates that token interaction requires not only
context aggregation, but also adaptive selection of propagated messages.
Meanwhile, removing the local branch causes smaller but consistent drops
of 0.11 and 0.07 percentage points, suggesting that local refinement
provides complementary spatial information.

We further compare different propagation structures within GSAP.
Replacing axial propagation with a full two-dimensional depthwise kernel
improves accuracy by 0.17 and 0.22 percentage points on CIFAR-10 and
CIFAR-100, respectively.
However, this improvement introduces substantially higher parameter costs.
Specifically, axial propagation requires only $2Ck$ spatial kernel
parameters, compared with $Ck^2$ for a full two-dimensional kernel,
reducing the propagation parameters by 86.7\% when $k=15$.
Although full two-dimensional propagation provides stronger spatial
interaction capability, axial propagation achieves a better balance
between representation capability and model complexity, making it a more
suitable design choice for GSAP. The corresponding ablation results are summarized in Fig.~\ref{fig:ablation}.

% \noindent
% \begin{minipage}{0.48\linewidth}
% \begin{equation}
% \label{eq:ottt_1}
% \hat{\mathbf{a}}^{l}[t]
% = \lambda \hat{\mathbf{a}}^{l}[t-1] + \mathbf{s}^{l}[t],
% \end{equation}
% \end{minipage}
% \hfill
% \begin{minipage}{0.48\linewidth}
% \begin{equation}
% \label{eq:ottt_2}
% \widehat{\nabla}_{\mathbf{W}^{l}}\mathcal{L}_{t}
% = \mathbf{g}^{l+1}[t]
% \left(\hat{\mathbf{a}}^{l}[t]\right)^{\top}.
% \end{equation}
% \end{minipage}
\subsection{GSAP for Online Learning}

Beyond inference-time representation learning, token interaction also
influences the temporal propagation of neural activity and learning
signals during optimization.
Unlike attention-based interaction that relies on instantaneous token
affinity, GSAP introduces structured propagation across tokens, which may
provide more consistent information flow under online learning settings~\citep{kaiser2020synaptic}.

We investigate this issue within an OTTT-style online learning framework~\citep{xiao2022online}, where gradients combine temporally accumulated neural traces with current-step learning signals.
In Spiking Attention, multiplicative terms involving sparse spike activations can suppress learning signals and cause gradient directions to fluctuate across time steps, hindering effective gradient accumulation. GSAP alleviates this bottleneck through structured spatial propagation that avoids reliance on pairwise spike matching. To isolate the contribution of propagation dynamics, we evaluate an ungated GSAP variant, which exhibits higher cross-step gradient cosine similarity and improved classification accuracy under OTTT. The corresponding analysis is provided in Appendix~\ref{ap:ottt}.

Figure~\ref{fig:online} compares the cross-step gradient cosine similarity
and final task performance between GSAP and Spiking Attention under OTTT.
The ungated GSAP variant achieves higher gradient similarity across time
steps, indicating more consistent learning signals during online updates. Furthermore, under the same experimental settings as the Spiking
Attention baseline, GSAP without receiver gating consistently improves
performance across the evaluated OTTT variants and tasks.
In particular, GSAP achieves a gain of more than 10 percentage points on
CIFAR-100 under OTTT-A (Fig.~\ref{fig:online}). These results demonstrate that propagation-based token interaction \textbf{not
only improves inference-time representation learning, but also provides
more stable information flow for online optimization in Spiking
Transformers.}
\begin{figure}[h]
    \centering
    \includegraphics[width=\textwidth]{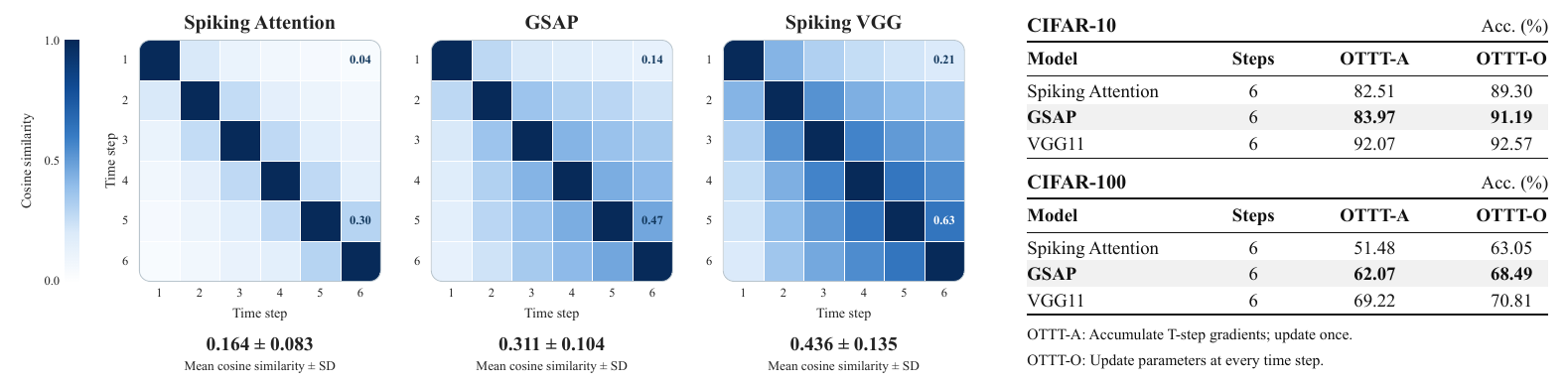}
\caption{\small
Gradient consistency and task performance under OTTT-based online learning. Spiking Attention struggles to maintain consistent gradient directions across time steps, hindering gradient accumulation and effective parameter updates. GSAP denotes the ungated variant used to isolate the effect of
propagation-based token interaction.
}
    \label{fig:online}
\end{figure}
\section{Conclusion}
In this work, we revisit token interaction in Spiking Transformers from a spike-native perspective and propose Gated Spike Axial Propagation (GSAP). Instead of relying on explicit pairwise token affinity, GSAP enables token communication through structured spike propagation and receiver-conditioned message selection. Experiments across image classification, event-based recognition, and semantic segmentation demonstrate the effectiveness and generality of GSAP. Ablation studies reveal that long-range propagation, adaptive message selection, and local refinement contribute to token interaction. Token-perturbation analyses show that long-range propagation can coexist with spatially localized sensitivity. This highlights the value of combining broad spatial communication with selective context utilization in spiking representations. Our findings suggest that Spiking Transformers may benefit from interaction mechanisms designed around the temporal and sparse properties of spiking computation rather than directly adapting conventional attention mechanisms.

\newpage
\section*{AI Use Statement}
Generative AI tools were used solely for language translation and final-stage polishing to improve the clarity, fluency, and readability of the manuscript. They were not used for research ideation, methodology design, experimental design, data analysis, result interpretation, or the formulation of scientific claims. The authors take full responsibility for the content of the manuscript.

\section*{Reproducibility Statement}
Detailed experimental configurations, including implementation and training settings, are provided in the Appendix. The complete experimental code necessary to reproduce the results is included in the supplementary material submitted with this paper.

\bibliography{iclr2027_conference}
\bibliographystyle{iclr2027_conference}

\newpage
\appendix
\section{Appendix}
\subsection{Neuron Encoding}
\label{ap:encoding}

Let $x(\mathbf{p})\in[0,1]$ denote the normalized input intensity or feature value at coordinate $\mathbf{p}$ and let $t\in\{1,\ldots,T\}$ denote the simulation step. We write $I_t$ for the input supplied by an encoder. Unlike internal binary neuron outputs, $I_t$ may be real-valued under direct or amplitude-weighted encoding.

\paragraph{Direct Encoding.}
The input value is directly injected into the network at every time step:
\begin{equation}
    I_t(\mathbf{p}) = x(\mathbf{p}),
    \qquad t = 1,\ldots,T.
    \label{eq:direct_encoding}
\end{equation}
Thus, the same input current is presented throughout the entire simulation
window, resulting in a temporally constant input sequence~\citep{rathi2021diet}.

\paragraph{Phase Encoding.}
We first quantize the normalized input into an 8-bit integer,
\begin{equation}
    v(\mathbf{p})
    =
    \left\lfloor 256\,x(\mathbf{p}) \right\rfloor,
    \label{eq:phase_quantization}
\end{equation}
and encode its binary representation across time as
\begin{equation}
I_t(\mathbf{p})
=
\begin{cases}
    2^{-(b+1)},
    & \text{if } v_{7-b}(\mathbf{p}) = 1, \\[3pt]
    0,
    & \text{otherwise},
\end{cases}
\qquad
b \equiv (t-1) \pmod{8},
\label{eq:phase_encoding}
\end{equation}
where $v_k(\mathbf{p})$ denotes the $k$-th bit of the 8-bit integer
$v(\mathbf{p})$, with $k=7$ corresponding to the most significant bit.
The encoder sequentially traverses the eight bit planes and assigns
larger weights to more significant bits~\citep{kim2018deep}.

\paragraph{Rate Encoding.}
For rate encoding, spikes are sampled independently at each time step
according to a Bernoulli distribution:
\begin{equation}
    I_t(\mathbf{p})
    \sim
    \operatorname{Bernoulli}\!\left(x(\mathbf{p})\right),
    \qquad
    \mathbb{E}\!\left[I_t(\mathbf{p})\right]
    =
    x(\mathbf{p}),
    \qquad
    t = 1,\ldots,T.
    \label{eq:rate_encoding}
\end{equation}
Therefore, the expected firing rate over the simulation window is
proportional to the magnitude of the input feature~\citep{rueckauer2017conversion}.

\paragraph{Time-to-First-Spike (TTFS) Encoding.}
For TTFS encoding, each input neuron emits a single spike, whose latency
is inversely related to the input magnitude. We define the firing time as
\begin{equation}
    t^{*}(\mathbf{p})
    =
    1
    +
    \left\lfloor
        \bigl(1-x(\mathbf{p})\bigr)(T-1)
    \right\rfloor .
    \label{eq:ttfs_time}
\end{equation}
The corresponding spike sequence is
\begin{equation}
I_t(\mathbf{p})
=
\begin{cases}
    \displaystyle \frac{1}{t^{*}(\mathbf{p})},
    & t = t^{*}(\mathbf{p}), \\[7pt]
    0,
    & \text{otherwise}.
\end{cases}
\label{eq:ttfs_encoding}
\end{equation}
Hence, larger input values produce earlier spikes, whereas smaller values
produce later spikes. The factor $1/t^{*}(\mathbf{p})$ can be used to
modulate the spike amplitude; alternatively, a binary spike with unit
amplitude may be adopted~\citep{mostafa2017supervised}.

\subsection{Experiments on Encoding Methods}
\label{ap:neuron encoding}
\vspace{-5mm}
\begin{table}[h]
    \centering
        \caption{\small Performance of baselines \& GSAP on different neuron encodings.}
        \resizebox{0.35\linewidth}{!}{%
            \begin{tabular}{lcccc}
        \toprule
        \textbf{Model} & \textbf{Direct} & \textbf{Phase} & \textbf{Rate} & \textbf{TTFS} \\
        \midrule
        Spikformer  & 95.51 & 82.63 & 82.68 & 81.87 \\
        SDT         & 95.60 & 85.33 & 84.06 & 84.52 \\
        TIM         & 94.20 & 81.43 & 81.48 & 80.66 \\
        QKFormer    & 96.18 & 87.76 & 83.77 & 84.69 \\
        TEFormer    & 96.24 & 89.92 & 84.74 & 87.46 \\
        \midrule
        \textbf{GSAP (ours)} & 95.95 & \textbf{92.51} & \textbf{85.46} & \textbf{90.96} \\
        \bottomrule
        \end{tabular}
        }
\end{table}

We compare direct, phase, rate, and TTFS input encoding to assess GSAP across different spiking representations. GSAP reaches 92.51\%, 85.46\%, and 90.96\% accuracy under the three alternative encodings, respectively, outperforming the compared models in each setting. TEFormer, the strongest alternative-encoding baseline in the table, reaches 89.92\%, 84.74\%, and 87.46\%. Under direct encoding, GSAP reaches 95.95\%, below TEFormer at 96.24\%. Thus, GSAP performs well across the tested alternative encodings even relative to models with stronger direct-encoding results, supporting its applicability beyond direct encoding.

\newpage
\section{GSAP Global Modeling Capability}
\label{ap:effectiveness}
\begin{figure}[h]
    \centering
    \includegraphics[width=0.80\linewidth]{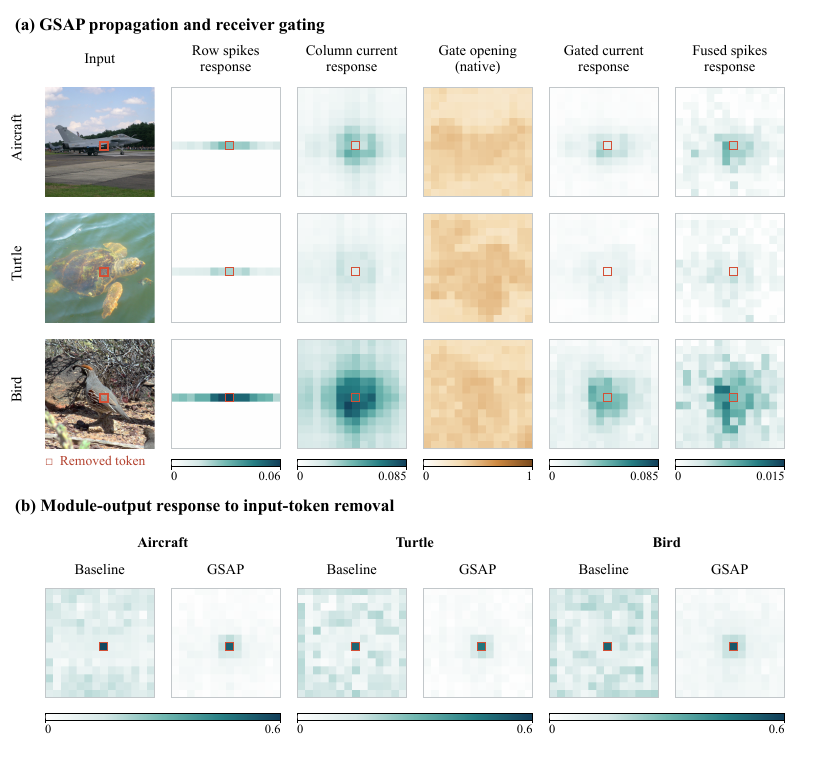}
    \caption{Token perturbation analysis of GSAP on the trained ImageNet-1K
    model.
    (a) Removing a central feature token reveals axial propagation and
    receiver-conditioned message selection.
    (b) Interaction responses after removing an input token, where the GSAP
    gate responds naturally.
    Response magnitude indicates perturbation sensitivity.}
    \label{fig:effectiveness}
\end{figure}

\paragraph{Global spatial propagation and receiver gating.}
\label{sec:mechanism}
To understand how GSAP performs token interaction beyond explicit pairwise affinity modeling, we analyze the propagation behavior of a trained ImageNet-1K Spikingformer-8-768 model with $T=4$. Input images are resized and center-cropped to $224\times224$ with a crop ratio of $1.0$, followed by ImageNet normalization. We examine the fourth block, which operates on a $14\times14$ token grid, in evaluation mode with spiking-neuron states reset before each reference or perturbed replay. To isolate the global propagation pathway, we zero the central token at the output of the feature spiking neuron across all channels and time steps, while fixing the receiver gate to its unperturbed value and leaving the local and identity branches unchanged. We visualize the absolute difference between the reference and perturbed activations, averaged over channels and time. As shown in 
\begin{equation}
\label{eq:gsap_receptive_field}
\begin{aligned}
    \text{Since} \quad & |j-v| \leq W-1 \leq \frac{k-1}{2} \quad \text{and} \quad |i-u| \leq H-1 \leq \frac{k-1}{2}, \\
    \implies \quad & (i,j) \rightarrow (i,v) \rightarrow (u,v) \quad \text{(valid structural path)}, \\
    \therefore \quad & \mathcal{R}_{\mathrm{struct}}(M^{t}_{u,v}) = \bigcup_{i=1}^{H}\{(i,j): 1 \leq j \leq W\} = \{1,\ldots,H\} \times \{1,\ldots,W\}.
\end{aligned}
\end{equation}
Fig.~\ref{fig:effectiveness}(a), the perturbation first propagates along the source row and subsequently extends across rows through column propagation, producing a two-dimensional response within the same simulation step. Meanwhile, the receiver gate selectively attenuates the propagated current before spike fusion. This behavior directly reflects the axial propagation pathway described in Eq.~\ref{eq:gsap_receptive_field}, providing empirical evidence that GSAP establishes structured long-range communication while retaining receiver-dependent control over the transmitted signal.

\paragraph{Spatial sensitivity to token removal.}
We further examine whether such global propagation leads to spatially indiscriminate responses by comparing the trained baseline and GSAP models under the same token-removal protocol. For each model, we zero the central input token of the fourth interaction block across all channels and time steps, and measure the resulting change in the interaction-branch output before the outer residual addition. The absolute response is averaged over channels and time, with each model receiving its own intermediate features; unlike the preceding pathway-isolation experiment, the GSAP receiver gate is allowed to respond naturally to the perturbed input. All response maps in Fig.~\ref{fig:effectiveness}(b) share a common absolute color scale. In the displayed examples, the baseline exhibits broadly distributed output changes, whereas GSAP produces a more spatially concentrated response around the removed token, with weaker changes at distant positions. Together with the propagation visualization in panel~(a), this result shows that GSAP combines long-range information propagation with localized sensitivity to input perturbations: information can be transmitted over distant spatial positions without inducing uniformly strong influence across the token grid. The response maps characterize perturbation sensitivity rather than absolute activation magnitude, and their absolute scales also depend on the intermediate feature distributions. The complete intervention protocol is provided in Appendix~\ref{ap:effectiveness}.

\newpage
\section{Experiment}
\subsection{ImageNet}
\label{ap:imagenet}
\begin{table*}[h]
\caption{
Architecture specifications of the eight-block GSAP models on
ImageNet.
S-ST denotes a Spikingformer-style single-stage backbone,
and H-ST denotes a QKFormer-style hierarchical backbone.
$C_S$ and $C_H$ are their final channel dimensions.
S-ST applies all eight Transformer blocks at a fixed resolution,
whereas H-ST distributes them across three resolutions with
depths $(1,2,5)$.
Axial kernel sizes are specified for $H=W=224$ and follow
$k=2\max(H_s,W_s)-1$ at each feature resolution.
PW and DW denote pointwise and depthwise convolutions,
respectively; SN denotes a LIF neuron, with the subscript
indicating a firing threshold of $0.5$.
Normalization and neuron placements in the tokenizer, patch
embedding, and MLP follow the respective backbone implementations.
}
\centering
\begingroup
\renewcommand{\arraystretch}{1.35}
\setlength{\tabcolsep}{7pt}
\setlength{\arrayrulewidth}{0.45pt}

\resizebox{\textwidth}{!}{%
\begin{tabular}{|c|c|c|l|c|c|}
\hline
\textbf{Stage}
& \textbf{Spatial resolution}
& \multicolumn{2}{c|}{\textbf{Layer specification}}
& \textbf{GSAP / S-ST}
& \textbf{GSAP / H-ST} \\
\hline
\multirow{3}{*}{Tokenizer}
& \multirow{3}{*}{\makecell{$H\times W$\\$\downarrow$\\Feature grid}}
& Convolutions
& Conv $3\times3$, stride 1
& \makecell{$C_S/8,\ C_S/4,\ C_S/2,$\\$C_S,\ C_S$}
& $C_H/8,\ C_H/4,\ C_H/4$ \\
\cline{3-6}

& & Downsampling
& MaxPool $3\times3$, stride 2
& $\times4$
& $\times2$ \\
\cline{3-6}

& & Output
& Resolution; channel dimension
& $\dfrac{H}{16}\times\dfrac{W}{16};\ C_S$
& $\dfrac{H}{4}\times\dfrac{W}{4};\ C_H/4$ \\
\hline
\multirow{2}{*}{Fine}
& \multirow{2}{*}{$\dfrac{H}{4}\times\dfrac{W}{4}$}
& Blocks
& GSAP Transformer block
& ---
& $\times1$, dim $C_H/4$ \\
\cline{3-6}

& & Axial kernels
& DWConv $1\times k$; DWConv $k\times1$
& ---
& $k=111$ \\
\hline
\multirow{3}{*}{Intermediate}
& \multirow{3}{*}{$\dfrac{H}{8}\times\dfrac{W}{8}$}
& Downsampling
& Residual patch embedding, stride 2
& ---
& $C_H/4\rightarrow C_H/2$ \\
\cline{3-6}

& & Blocks
& GSAP Transformer block
& ---
& $\times2$, dim $C_H/2$ \\
\cline{3-6}

& & Axial kernels
& DWConv $1\times k$; DWConv $k\times1$
& ---
& $k=55$ \\
\hline
\multirow{3}{*}{Coarse}
& \multirow{3}{*}{$\dfrac{H}{16}\times\dfrac{W}{16}$}
& Downsampling
& Residual patch embedding, stride 2
& ---
& $C_H/2\rightarrow C_H$ \\
\cline{3-6}

& & Blocks
& GSAP Transformer block
& $\times8$, dim $C_S$
& $\times5$, dim $C_H$ \\
\cline{3-6}

& & Axial kernels
& DWConv $1\times k$; DWConv $k\times1$
& $k=27$
& $k=27$ \\
\hline
\multirow{2}{*}{Head}
& \multirow{2}{*}{$1\times1$}
& Aggregation
& Spatial and temporal averaging
& \multicolumn{2}{c|}{Global average pooling; temporal mean} \\
\cline{3-6}

& & Classifier
& Fully connected layer
& $C_S\rightarrow1000$
& $C_H\rightarrow1000$ \\
\hline

\multicolumn{6}{|c|}{\textit{\textbf{Block specifications}}} \\
\hline

\multirow{2}{*}{\makecell{Residual patch\\embedding}}
& \multicolumn{2}{c|}{Main pathway}
& \multicolumn{3}{l|}{
  Conv $3\times3$
  $\rightarrow$ MaxPool, stride 2
  $\rightarrow$ Conv $3\times3$
} \\
\cline{2-6}

& \multicolumn{2}{c|}{Shortcut}
& \multicolumn{3}{l|}{
  Conv $1\times1$, stride 2; additive fusion
} \\
\hline

\multirow{7}{*}{GSAP}
& \multicolumn{2}{c|}{Input encoding}
& \multicolumn{3}{l|}{
  LIF neuron: $\mathbf{S}=\operatorname{SN}(\mathbf{X})$
} \\
\cline{2-6}

& \multicolumn{2}{c|}{Local pathway}
& \multicolumn{3}{l|}{
  DWConv $3\times3$ + BN $\rightarrow \mathbf{L}$
} \\
\cline{2-6}

& \multicolumn{2}{c|}{Feature projection}
& \multicolumn{3}{l|}{
  PWConv $1\times1$ + BN + SN $\rightarrow \mathbf{F}$
} \\
\cline{2-6}

& \multicolumn{2}{c|}{Axial propagation}
& \multicolumn{3}{l|}{\makecell[l]{
  $\mathbf{H}=\operatorname{SN}
  \bigl(\mathbf{F}+\operatorname{BN}
  (\operatorname{DWConv}_{1\times k}(\mathbf{F}))\bigr)$\\
  $\mathbf{M}=\operatorname{BN}
  (\operatorname{DWConv}_{k\times1}(\mathbf{H}))$
}} \\
\cline{2-6}

& \multicolumn{2}{c|}{Receiver gate}
& \multicolumn{3}{l|}{
  PWConv $1\times1$ + BN + SN$_{0.5}$
  $\rightarrow \mathbf{G}$
} \\
\cline{2-6}

& \multicolumn{2}{c|}{Gated fusion}
& \multicolumn{3}{l|}{
  $\mathbf{A}=\operatorname{SN}_{0.5}
  \bigl(\operatorname{BN}
  (\mathbf{S}+\mathbf{L}+\mathbf{M}\odot\mathbf{G})\bigr)$
} \\
\cline{2-6}

& \multicolumn{2}{c|}{Output projection}
& \multicolumn{3}{l|}{
  PWConv $1\times1$ + BN; channel dimension preserved
} \\
\hline

\multirow{2}{*}{\makecell{GSAP Transformer\\block}}
& \multicolumn{2}{c|}{Spatial interaction}
& \multicolumn{3}{l|}{
  $\mathbf{Y}=\mathbf{X}+\operatorname{GSAP}(\mathbf{X})$
} \\
\cline{2-6}

& \multicolumn{2}{c|}{Channel MLP}
& \multicolumn{3}{l|}{
  Conv $1\times1\times2$, expansion ratio 4;
  $\mathbf{Out}=\mathbf{Y}+\operatorname{MLP}(\mathbf{Y})$
} \\
\hline
\end{tabular}%
}

\label{tab:gsap_imagenet_architecture}
\endgroup
\end{table*}
\paragraph{ImageNet training configurations.}
We use RGB inputs of size $3\times224\times224$ and $T=4$ time steps.
Images are normalized using channel-wise means
$(0.485,0.456,0.406)$ and standard deviations
$(0.229,0.224,0.225)$, with bicubic interpolation for resizing.
Both the baseline and GSAP models are trained with AdamW,
a weight decay of $0.05$, and a per-GPU batch size of $24$.
The training schedule comprises $300$ epochs followed by
a $10$-epoch cooldown, with $20$ warmup epochs starting
from a learning rate of $10^{-6}$ and cosine decay
to a minimum learning rate of $10^{-5}$.
The baseline uses a peak learning rate of $5\times10^{-4}$.
For GSAP, this value is scaled linearly by $B/288$,
where $B$ denotes the global training batch size.
Training augmentation includes random horizontal flipping
with probability $0.5$, RandAugment
(\texttt{rand-m9-mstd0.5-inc1}), Mixup with $\alpha=0.8$,
CutMix with $\alpha=1.0$, and random erasing with probability $0.25$.
Mixup and CutMix are selected with equal probability.
We use label smoothing of $0.1$ and automatic mixed-precision training.

\newpage

\newpage
\subsection{Segmentation}
\label{ap:seg}
\begin{table*}[h]
\centering
\caption{
Architecture specifications of the GSAP segmentation models on ADE20K.
Both backbone adaptations use a shared multi-scale layout, with
GSAP replacing all eight attention modules in the final two stages.
$P_1$--$P_4$ denote the backbone features passed to the FPN neck
after temporal averaging.
The axial kernels are configured for $512\times512$ training crops,
corresponding to a $32\times32$ feature grid in both GSAP stages.
PW and DW denote pointwise and depthwise convolutions;
BN and SN denote batch normalization and LIF spiking neurons.
The subscript $0.5$ indicates the firing threshold.
}
\begingroup
\renewcommand{\arraystretch}{1.3}
\setlength{\tabcolsep}{6pt}
\setlength{\arrayrulewidth}{0.45pt}

\resizebox{\textwidth}{!}{%
\begin{tabular}{|c|c|c|l|c|c|}
\hline
\textbf{Stage}
& \textbf{Spatial resolution}
& \multicolumn{2}{c|}{\textbf{Layer specification}}
& \makecell{\textbf{GSAP}\\Spikingformer-based}
& \makecell{\textbf{GSAP}\\QKFormer-based} \\
\hline
\multirow{2}{*}{1}
& \multirow{2}{*}{$\dfrac{H}{2}\times\dfrac{W}{2}$}
& Downsampling
& Conv $7\times7$, stride 2, dim 32
& \multicolumn{2}{c|}{Shared configuration} \\
\cline{3-6}

& & Blocks
& Conv-based SNN block
& $\times1\rightarrow P_1$
& $\times1\rightarrow P_1$ \\
\hline
\multirow{2}{*}{2}
& \multirow{2}{*}{$\dfrac{H}{4}\times\dfrac{W}{4}$}
& Downsampling
& Conv $3\times3$, stride 2, dim 64
& \multicolumn{2}{c|}{Shared configuration} \\
\cline{3-6}

& & Blocks
& Conv-based SNN block
& $\times1\rightarrow P_2$
& $\times1\rightarrow P_2$ \\
\hline
\multirow{2}{*}{3}
& \multirow{2}{*}{$\dfrac{H}{8}\times\dfrac{W}{8}$}
& Downsampling
& Conv $3\times3$, stride 2, dim 128
& \multicolumn{2}{c|}{Shared configuration} \\
\cline{3-6}

& & Blocks
& Conv-based SNN block
& $\times2\rightarrow P_3$
& $\times2\rightarrow P_3$ \\
\hline
\multirow{2}{*}{4}
& \multirow{2}{*}{$\dfrac{H}{16}\times\dfrac{W}{16}$}
& Downsampling
& Conv $3\times3$, stride 2, dim 256
& \multicolumn{2}{c|}{Shared configuration} \\
\cline{3-6}

& & Blocks
& GSAP Transformer block
& GSAP $\times6$
& GSAP $\times6$ \\
\hline
\multirow{2}{*}{5}
& \multirow{2}{*}{$\dfrac{H}{16}\times\dfrac{W}{16}$}
& Channel expansion
& Conv $3\times3$, stride 1, dim 360
& \multicolumn{2}{c|}{Shared configuration} \\
\cline{3-6}

& & Blocks
& GSAP Transformer block
& GSAP $\times2\rightarrow P_4$
& GSAP $\times2\rightarrow P_4$ \\
\hline
\multicolumn{6}{|c|}{\textit{\textbf{Block specifications}}} \\
\hline

\multirow{2}{*}{\makecell{Conv-based\\SNN block}}
& \multicolumn{2}{c|}{SepConv}
& \multicolumn{3}{l|}{
  PW $1\times1$ (ratio 2)
  $\rightarrow$ DW $7\times7$
  $\rightarrow$ PW $1\times1$; membrane shortcut
} \\
\cline{2-6}

& \multicolumn{2}{c|}{Channel convolution}
& \multicolumn{3}{l|}{
  Conv $3\times3\times2$; expansion ratio 4;
  membrane shortcut
} \\
\hline

\multirow{6}{*}{\makecell{GSAP Transformer\\block}}
& \multicolumn{2}{c|}{Local pathway}
& \multicolumn{3}{l|}{
  DW $3\times3$ + BN, applied to input spikes
} \\
\cline{2-6}

& \multicolumn{2}{c|}{Propagation pathway}
& \multicolumn{3}{l|}{\makecell[l]{
  PW $1\times1$ + BN + SN
  $\rightarrow$ DW $1\times63$ + BN\\
  $\rightarrow$ feature shortcut + SN
  $\rightarrow$ DW $63\times1$ + BN
}} \\
\cline{2-6}

& \multicolumn{2}{c|}{Receiver gate}
& \multicolumn{3}{l|}{
  PW $1\times1$ + BN + SN$_{0.5}$;
  element-wise gating of propagated features
} \\
\cline{2-6}

& \multicolumn{2}{c|}{Fusion and projection}
& \multicolumn{3}{l|}{\makecell[l]{
  Input spikes + local features + gated propagated features\\
  $\rightarrow$ BN + SN$_{0.5}$
  $\rightarrow$ PW $1\times1$ + BN
}} \\
\cline{2-6}

& \multicolumn{2}{c|}{Channel MLP}
& \multicolumn{3}{l|}{
  Conv $1\times1\times2$; expansion ratio 4
} \\
\cline{2-6}

& \multicolumn{2}{c|}{Outer residuals}
& \multicolumn{3}{l|}{
  $\mathbf{Y}=\mathbf{X}+\operatorname{GSAP}(\mathbf{X}),\quad
   \mathbf{Z}=\mathbf{Y}+\operatorname{MLP}(\mathbf{Y})$
} \\
\hline
\multicolumn{6}{|c|}{\textit{\textbf{Feature pyramid and segmentation head}}} \\
\hline

\multirow{2}{*}{Feature interface}
& \multicolumn{2}{c|}{Backbone outputs}
& \multicolumn{3}{l|}{
  $P_1$--$P_4$: channels $(32,64,128,360)$;
  spatial strides $(2,4,8,16)$
} \\
\cline{2-6}

& \multicolumn{2}{c|}{Temporal aggregation}
& \multicolumn{3}{l|}{
  Temporal mean of each feature map over $T=4$ steps
} \\
\hline

\multirow{2}{*}{FPN}
& \multicolumn{2}{c|}{Lateral projection}
& \multicolumn{3}{l|}{
  $1\times1$ convolution; 128 output channels per level
} \\
\cline{2-6}

& \multicolumn{2}{c|}{Pyramid fusion}
& \multicolumn{3}{l|}{
  Top-down upsampling and addition;
  $3\times3$ output convolutions
} \\
\hline

\multirow{2}{*}{Decode head}
& \multicolumn{2}{c|}{Multi-scale fusion}
& \multicolumn{3}{l|}{
  128-channel scale branches; bilinear resizing and summation
} \\
\cline{2-6}

& \multicolumn{2}{c|}{Prediction}
& \multicolumn{3}{l|}{
  $1\times1$ classifier, 150 classes;
  resize logits to the image resolution
} \\
\hline
\end{tabular}%
}
\label{tab:gsap_segmentation_architecture}
\endgroup
\end{table*}

\paragraph{Segmentation configurations.}
Table~\ref{tab:gsap_segmentation_architecture} details the
GSAP configurations for semantic segmentation on ADE20K.
We adapt the Spikingformer-based and QKFormer-based backbones
to a shared multi-scale layout using $512\times512$ training
crops and $T=4$ time steps.
The first three stages contain $(1,1,2)$ convolutional SNN
blocks with channel widths $(32,64,128)$.
The final two stages operate at an output stride of $16$
and contain six and two GSAP Transformer blocks with channel
widths of $256$ and $360$, respectively.
Each Transformer block uses an MLP expansion ratio of $4$.
GSAP replaces all eight attention modules while retaining
the corresponding backbone's convolutional stages, channel
transitions, MLPs, and residual connections.
For the resulting $32\times32$ training feature grids,
each GSAP module uses depthwise axial kernels of sizes
$1\times63$ and $63\times1$.

The backbone outputs four feature maps with channel widths
$(32,64,128,360)$ at spatial strides $(2,4,8,16)$.
These features are averaged over time and passed to a
128-channel FPN neck and a 128-channel segmentation head
for 150-class prediction.
Both models are trained from scratch, and evaluation uses
whole-image inference.
The axial kernel sizes remain fixed during inference;
their spatial coverage therefore does not automatically
expand for images larger than the training crops.

\newpage
\paragraph{Visualization}
\begin{figure}[h]
    \centering
    \includegraphics[width=1\linewidth]{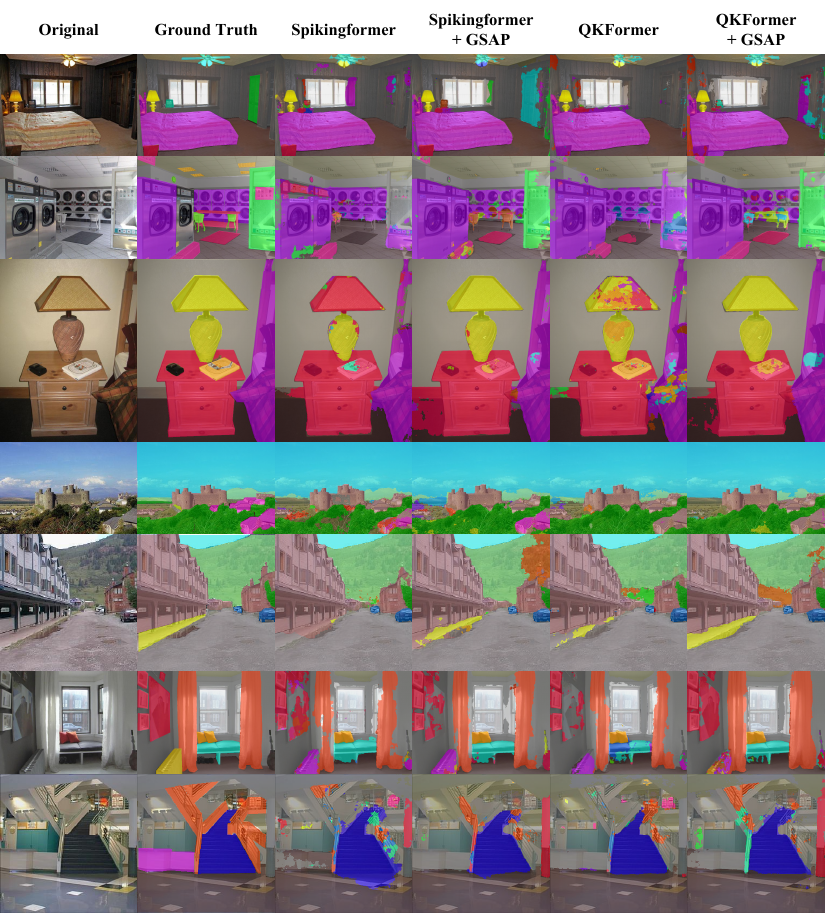}
    \caption{Selected qualitative semantic segmentation results on the ADE20K validation set. Columns show the input image, ground truth, Spikingformer, Spikingformer with GSAP, QKFormer, and QKFormer with GSAP. Rows show a bedroom, a laundromat, a lamp, a castle, a street scene, a curtained room, and an indoor staircase.}
    \label{fig:seg}
\end{figure}

Figure~\ref{fig:seg} illustrates local improvements in semantic region consistency with GSAP across two backbones. In the lamp example, GSAP corrects the lampshade prediction in Spikingformer and reduces fragmented labels in QKFormer. In the laundromat, both variants recover more of the rightmost machine with the ground-truth class. The outdoor examples show more coherent vegetation labeling around the castle, particularly for Spikingformer, and more complete sidewalk predictions in the street scene, particularly for QKFormer. The added indoor examples show more consistent curtain labeling and recovery of parts of the staircase structure. These examples were selected to highlight improvements; errors remain in some regions, such as the bedroom doorway and street vegetation.

\newpage
\section{Online Learning Adaption}
\label{ap:ottt}
\paragraph{SNNs Online Learning}
In online learning for SNNs, gradients are typically accumulated during single-step forward propagation, followed by periodic backward updates. Taking OTTT as an example, neuronal dynamics accumulate temporal traces $\hat{\mathbf{a}}^{l}[t]$ (Eq.~\ref{eq:ottt_1}), while the learning signal is propagated through the computational graph to the neurons and combined with these traces to construct the full gradients of the corresponding weights (Eq.~\ref{eq:ottt_2}). This learning paradigm preserves both computational efficiency and strict causality~\citep{xiao2022online}.
\noindent
\begin{minipage}{0.48\linewidth}
\begin{equation}
\label{eq:ottt_1}
\hat{\mathbf{a}}^{l}[t]
= \lambda \hat{\mathbf{a}}^{l}[t-1] + \mathbf{s}^{l}[t],
\end{equation}
\end{minipage}
\hfill
\begin{minipage}{0.48\linewidth}
\begin{equation}
\label{eq:ottt_2}
\nabla_{\mathbf{W}^{l}}\mathcal{L}_{t}
= \mathbf{g}^{l+1}[t]
\left(\hat{\mathbf{a}}^{l}[t]\right)^{\top}.
\end{equation}
\end{minipage}
vfsd
\paragraph{Spiking Transformer Adaptation.}
Existing online learning algorithms have been developed primarily for spike-based CNNs, and we find that they are difficult to adapt effectively to Spiking Transformers.
According to Eqs.~\ref{eq:ottt_1} and~\ref{eq:ottt_2}, weight gradients are constructed from temporal traces and learning signals propagated through the chain rule. However, sparse QKV activity can suppress components of the learning signal, as indicated by the \textcolor{red}{red term} in Eq.~\ref{eq:ottt_st}.

\begin{equation}
\nabla_{\mathbf{W}_{Q}^{l}}\mathcal{L}_{t}
=
\mathbf{g}_{Q}^{l+1}[t]
\left(\hat{\mathbf{a}}^{l}[t]\right)^{\top},
\mathbf{g}_{Q}^{l+1}[t]
=
\phi'\!\left(
\mathbf{U}_{Q}^{l+1}[t]-\vartheta
\right)
\odot
\textcolor{red}{
\left(
s\mathbf{D}_{t}\mathbf{V}_{t}^{\top}\mathbf{K}_{t}
\right)
}.
\label{eq:ottt_st}
\end{equation}

Sparse learning signals can limit gradient contributions within each time step, while weak alignment across steps may further reduce the effectiveness of gradient accumulation. As shown in Fig.~\ref{fig:online}, under OTTT-A, Spiking Attention exhibits lower cross-step gradient cosine similarity than the ungated GSAP variant.

We remove the receiver gate from GSAP and evaluate the resulting ungated variant on the same benchmarks under identical experimental settings. Across both OTTT variants and evaluated tasks, this variant consistently improves accuracy over the Spiking Attention baseline. In particular, under OTTT-A on CIFAR-100, it improves accuracy by more than 10 percentage points (Fig.~\ref{fig:online}). These results support propagation-based token interaction as a promising approach to improving online learning in Spiking Transformers.

\paragraph{OTTT Configuration}
We evaluate the spiking-attention baseline and GSAP on
CIFAR-10 and CIFAR-100 using OTTT-A and OTTT-O.
Both variants compute gradients at each time step.
OTTT-A accumulates these gradients over the complete sequence
and updates the parameters once, whereas OTTT-O updates
the parameters immediately after each time step.
Thus, the two variants perform one and $T$ parameter updates
per input sequence, respectively.

For both datasets, we use $3\times32\times32$ inputs and
present each image for $T=6$ time steps.
Training runs for $200$ epochs with a batch size of $128$,
using SGD with momentum $0.9$ and zero weight decay.
The learning rate starts at $0.1$ and follows a cosine
annealing schedule with a period of $200$ epochs and
a minimum value of zero, without warmup.
Training augmentation consists of random cropping with
four-pixel padding, Cutout, and random horizontal flipping.
Both datasets use channel-wise normalization with means
$(0.4914,0.4822,0.4465)$ and standard deviations
$(0.2023,0.1994,0.2010)$.
The LIF membrane time constant is $\tau=2$, and the
sigmoid surrogate gradient uses $\alpha=4$.
At each time step, the loss is
$[0.95\mathcal{L}_{\mathrm{CE}}+
0.05\mathcal{L}_{\mathrm{MSE}}]/T$,
where the MSE term compares the model output with
the one-hot class target.
Neuron states are reset at the start of each input sequence,
and predictions are obtained by summing the outputs
over all time steps.
We use a random seed of $2022$.

\newpage
\section{Complexity and Energy Analysis}
\label{ap:complexity}
\paragraph{Configurations and counting.}
We analyze the full local implementations using a common effective activity rate of $\rho=0.1$ to isolate architectural cost. CIFAR-10 and TinyImageNet use inputs of $32\times32$ and $64\times64$, respectively, with $T=4$, four blocks, a final width of 384, and an MLP expansion ratio of four. Their QK backbones use stage widths of 96/192/384 and block counts of 1/1/2. CIFAR10-DVS uses two-channel $128\times128$ event-count frames and $T=16$. The single-stage DVS backbones use two blocks and 256 channels. For the two-stage DVS QK backbone, we report both its native MLP ratio of one and a ratio-four structural control, keeping the ratio unchanged within each baseline--GSAP pair. These are architecture-level comparisons without checkpoint accuracy evaluation.

The FLOP counts in Tab.~\ref{tab:complexity} include convolutions, linear projections, and executed matrix products over the complete temporal sequence, with two FLOPs per multiply--accumulate (MAC). Parameter counts include all registered parameters. We obtain operation counts from full-size forward executions and verify them against an independent operator counter and analytical GSAP counts. The accounting excludes neuron-state updates, normalization, pooling, bias additions, general elementwise operations, memory access, and communication.

\paragraph{Normalized synaptic energy.}
Let $A$ denote the number of MACs in the input encoder, non-spiking classification heads, and non-spiking temporal mixing. Let $C$ denote the remaining convolution, linear, and attention-matrix MAC capacity, and $R$ the accumulation capacity of QK token reduction and SDSA reduction. All counts already include $T$. Following the arithmetic-cost convention in QKFormer~\cite{zhou2024qkformer}, we use $E_{\mathrm{MAC}}=4.6$~pJ and $E_{\mathrm{AC}}=0.9$~pJ and define
\begin{equation}
 E_{\mathrm{norm}}(\rho)
 = \bigl[4.6A + 0.9\rho(C+R)\bigr]\times10^{-9}\ \mathrm{mJ}.
 \label{eq:equal-rate-energy}
\end{equation}
The common rate is assigned to effective inputs of the normalized synaptic and reduction operations, including inputs after binary masking. For QK token attention and SDSA, each reduced input provides one accumulation opportunity, giving $R=TNC$ per block. This convention does not assume independence between the underlying spike tensors. The input convolution is charged at every executed time step. Event-count frames and temporally or spatially averaged classifier inputs retain their MAC cost; the SDT head is normalized as AC because it is preceded by a spiking neuron. GSAP computes its axial message before applying its gate, so the gate introduces no additional discount on axial convolution cost.

Equation~\eqref{eq:equal-rate-energy} is a controlled structural normalization. Equal marginal LIF firing rates alone do not determine joint mask activity or the cost of multi-valued residual inputs. In particular, some Spikformer and QKFormer projections consume sums of spikes. Their implementation may require dense MACs or multiple event accumulations, depending on the hardware mapping. Thus, the reported values are normalized synaptic energy rather than measured device energy or a complete energy prediction from firing rate alone.

\paragraph{Attention complexity.}
For one time step and one block with $N=HW$ tokens and $C$ channels, the implemented SSA evaluates $(QK^{\mathsf T})V$ and has MAC capacity
\begin{equation}
 \mathcal{C}_{\mathrm{SSA}}=4NC^2+2N^2C.
\end{equation}
GSAP uses three pointwise projections, a depthwise $3\times3$ local convolution, and two depthwise axial convolutions of length $k$, yielding
\begin{equation}
 \mathcal{C}_{\mathrm{GSAP}}=3NC^2+NC(9+2k).
\end{equation}
The current implementation sets $k=2S-1$ for a square $S\times S$ feature map. Consequently, its spatial propagation scales as $O(CN^{3/2})$, while its intermediate feature representation scales as $O(NC)$ and avoids an explicit $N\times N$ attention matrix. QK token attention already uses $3NC^2$ projection capacity and an $O(NC)$ reduction, so replacing these early-stage modules adds spatial mixing cost that offsets part of the savings from replacing later SSA blocks.

\paragraph{Comparison.}
On the Spikingformer backbone, GSAP reduces full-network FLOPs by 4.98\%, 8.35\%, and 2.99\% on CIFAR-10, TinyImageNet, and CIFAR10-DVS, respectively. At $\rho=0.1$, the corresponding normalized energy reductions are 4.65\%, 7.81\%, and 1.33\%. On QKFormer, the FLOP reductions are 0.43\%, 0.52\%, and 0.35\%, with normalized energy reductions of 0.42\%, 0.51\%, and 0.24\%; the DVS comparison uses the native MLP ratio of one. These results quantify the cost of substituting GSAP within each backbone under the stated normalization.

% Preamble: \usepackage{booktabs}
\begin{table*}[t]
\centering
\small
\caption{Complexity and normalized energy at $\rho=0.1$.
Arrows indicate baseline $\rightarrow$ GSAP.
Parentheses report reductions relative to the corresponding baseline.}
\label{tab:complexity}
\setlength{\tabcolsep}{5pt}
 \resizebox{1\linewidth}{!}{%
\begin{tabular}{llccc}
\toprule
\textbf{Dataset} & \textbf{Backbone} & \textbf{Params (M)} & \textbf{FLOPs (G)} & \textbf{Norm. energy (mJ)} \\
\midrule
\multirow{2}{*}{\textbf{CIFAR-10}}
& Spikingformer
& $9.3263 \rightarrow 8.8056$
& $7.4723 \rightarrow 7.1003$ ($\downarrow 4.98\%$)
& $0.3602 \rightarrow 0.3435$ ($\downarrow 4.65\%$) \\

& QKFormer
& $6.7062 \rightarrow 6.4748$
& $12.1406 \rightarrow 12.0886$ ($\downarrow 0.43\%$)
& $0.5703 \rightarrow 0.5679$ ($\downarrow 0.42\%$) \\
\midrule
\multirow{2}{*}{\textbf{TinyImageNet}}
& Spikingformer
& $9.3933 \rightarrow 8.9217$
& $31.0972 \rightarrow 28.5020$ ($\downarrow 8.35\%$)
& $1.4955 \rightarrow 1.3787$ ($\downarrow 7.81\%$) \\

& QKFormer
& $6.7794 \rightarrow 6.5971$
& $49.1663 \rightarrow 48.9084$ ($\downarrow 0.52\%$)
& $2.3088 \rightarrow 2.2970$ ($\downarrow 0.51\%$) \\
\midrule
\multirow{2}{*}{\textbf{CIFAR10-DVS}}
& Spikingformer
& $2.5687 \rightarrow 2.4586$
& $12.1132 \rightarrow 11.7514$ ($\downarrow 2.99\%$)
& $1.2261 \rightarrow 1.2098$ ($\downarrow 1.33\%$) \\

& QKFormer
& $1.5041 \rightarrow 1.4602$
& $16.2404 \rightarrow 16.1843$ ($\downarrow 0.35\%$)
& $1.0714 \rightarrow 1.0688$ ($\downarrow 0.24\%$) \\
\bottomrule
\end{tabular}}
\end{table*}

\end{document}